\documentclass{article} 
\PassOptionsToPackage{table}{xcolor}
\usepackage{iclr2025_conference,times}

\usepackage{amsmath,amsfonts,bm}

\def\eqref#1{equation~\ref{#1}}

\def\1{\bm{1}}

\DeclareMathAlphabet{\mathsfit}{\encodingdefault}{\sfdefault}{m}{sl}
\SetMathAlphabet{\mathsfit}{bold}{\encodingdefault}{\sfdefault}{bx}{n}

\usepackage[hyperfootnotes=false]{hyperref}
\usepackage{url}
\usepackage{graphicx}
\usepackage{booktabs} 
\usepackage{multirow} 
\usepackage{siunitx}  
\usepackage{ragged2e}
\usepackage[table]{xcolor}
\usepackage{marvosym}

\definecolor{mygray}{RGB}{220, 220, 220}
\newcommand\blfootnote[1]{%
  \begingroup
  \renewcommand\thefootnote{}\footnote{#1}%
  \addtocounter{footnote}{-1}%
  \endgroup
}

\title{FEVO: Financial Knowledge Expansion and Reasoning Evolution for Large Language Models}

\author{Bo Pang$^{1, \textrm{\Letter}}$, Yalu Ouyang$^{3, \textrm{\Cross}, \textrm{\Letter}}$, Hangfei Xu$^1$, Ziqi Jia$^{2, \textrm{\Cross}}$, Panpan Li$^1$, Shengzhao Wen$^1$, Lu Wang$^1$, Shiyong Li$^1$, Yanpeng Wang$^1$ \\
$^1$AI Cloud Group, Baidu $\quad ^2$Tsinghua University \\
$^3$University of California, San Diego \\
}

\iclrfinalcopy 
\begin{document}

\maketitle

\begin{abstract}
Advancements in reasoning for large language models (LLMs) have lead to significant performance improvements for LLMs in various fields such as mathematics and programming. However, research applying these advances to the financial domain, where considerable domain-specific knowledge is necessary to complete tasks, remains limited. To address this gap, we introduce \textbf{FEVO} (\textbf{F}inancial \textbf{Evo}lution), a multi-stage enhancement framework developed to enhance LLM performance in the financial domain. FEVO systemically enhances LLM performance by using continued pre-training (CPT) to \textit{expand financial domain knowledge}, supervised fine-tuning (SFT) to \textit{instill structured, elaborate reasoning patterns}, and reinforcement learning (RL) to \textit{further integrate the expanded financial domain knowledge with the learned structured reasoning}. To ensure effective and efficient training, we leverage frontier reasoning models and rule-based filtering to curate \textbf{FEVO-Train}, high-quality datasets specifically designed for the different post-training phases. Using our framework, we train the FEVO series of models\textemdash C32B, S32B, R32B\textemdash from Qwen2.5-32B and evaluate them on seven benchmarks to assess financial and general capabilities, with results showing that FEVO-R32B achieves state-of-the-art performance on five financial benchmarks against much larger models as well as specialist models. More significantly, FEVO-R32B demonstrates markedly better performance than FEVO-R32B-0 (trained from Qwen2.5-32B-Instruct using only RL), thus validating the effectiveness of \textit{financial domain knowledge expansion} and \textit{structured, logical reasoning distillation}.
\end{abstract}

\blfootnote{\textrm{\Cross} Completed during internship at Baidu}
\blfootnote{\textrm{\Letter} Emails: pangbo06@baidu.com, yaouyang@ucsd.edu}

\section{Introduction}

Recent studies in large language models (LLMs) have lead to widespread popularity of enhancing model capabilities via long Chain-Of-Thought (CoT) reasoning. Models such as OpenAI o1 \citep{openai-o1}, DeepSeek R1 \citep{deepseek-r1} have demonstrated sophisticated reasoning behaviors and remarkable performance on complex tasks. There have been many successful attempts \citep{wen2025light,hu2025openreasonerzeroopensourceapproach} to replicate this behavior on smaller models that have led to performances comparable to much larger models on tasks such as mathematics. In the financial domain, works such as Fin-R1 \citep{fin-r1} and DianJin-R1 \citep{dianjin-r1} have also validated the effectiveness of this reasoning paradigm.

However, the practical application of LLM reasoning models in financial scenarios still faces the following challenges: 

\begin{enumerate}
    \item Pretrained models display insufficient keyword expansion capability in the financial domain, which makes it difficult to cover the semantic requirements of professional terminology, emerging concepts, and complex logic when dealing with industry-specific scenarios.
    
    \item The low quality and poor verifiability of open-source data, with large amounts of data suffering from noise, latency, or unstructured issues and lacking a unique verifiable solution, directly affecting the reliability of model training and the credibility of reasoning results.

    \item In reinforcement learning (RL) training, ``reward hacking'' \citep{gao2022scalinglawsrewardmodel,everitt2021rewardtamperingproblemssolutions}\textemdash where models obtain illegal rewards by strategically bypassing true semantic understanding\textemdash often occurs, which elicit non-desirable model behaviors and undermine the effects of RL training.
\end{enumerate}

\begin{figure}[htbp]
\begin{center}
\includegraphics[width=\textwidth]{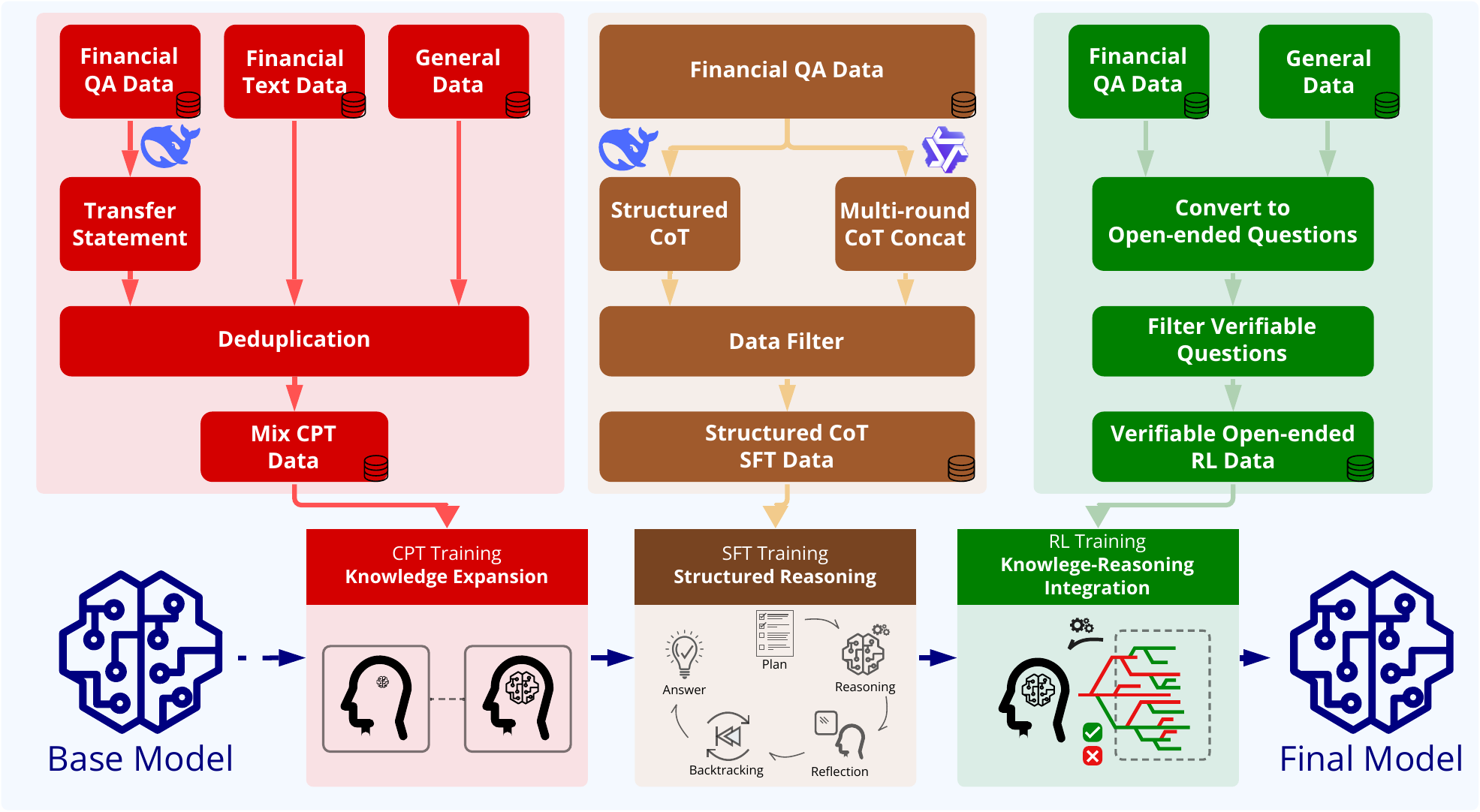}
\end{center}
\caption{Overall pipeline for \textbf{FEVO}, including dataset curation and training}
\label{fig:fevo-overall-pipeline}
\end{figure}

To address the above challenges, we designed \textbf{FEVO} (\textbf{F}inancial \textbf{Evo}lution), a framework to enhance LLM performance in the financial domain (as illustrated in Figure \ref{fig:fevo-overall-pipeline}). FEVO consists of three stages to progressively and systemically enhance LLM performance: continued pre-training (CPT) to \textbf{expand financial domain knowledge}, supervised fine-tuning (SFT) to \textbf{instill structured, elaborate reasoning patterns}, and reinforcement learning (RL) to \textbf{further integrate the expanded financial domain knowledge with the learned structured reasoning}. Primarily, FEVO addresses the issue of insufficient financial keyword expansion capability in pretrained models by using CPT to expand financial domain knowledge, then SFT and RL to guide models towards more effective utilization of their expanded knowledge during task-solving.

To obtain high-quality data for the different training phases, FEVO utilizes finely-grained filtering pipelines to elminate noise as well as corrupted content present in raw data. Specifically, the pipelines include answer-reference matching, reasoning chain validation, hyperlink elimination, image/table filtering, sub-question filtering and more. Additionally, to address the issue of ``reward hacking'' in the RL phase, we leverage DeepSeek-R1 to convert single-/multiple-choice questions into open-ended ones, vastly decreasing the possiblility of models ``guessing'' correct solutions and obtaining rewards without actually comprehending the question. Moreover, this process can effectively increase the amount of training data by expanding answer choices into standalone training samples. Using the filtering pipelines and conversion process, we curate \textbf{FEVO-Train-C, Train-S,} and \textbf{Train-R}\textemdash datasets tailored to each stage's training objectives.

We train the FEVO series of models\textemdash C32B, S32B, R32B\textemdash from Qwen2.5-32B and evaluate them on seven benchmarks to assess financial and general capabilities. Results show that \textbf{FEVO-R32B} achieves state-of-the-art performance on five financial benchmarks against much larger models and specialist models for the financial domain. We also train R32B-0 from Qwen2.5-Instruct using solely RL to gauge the effectiveness of our framework. Although FEVO-R32B-0 recorded the third-best overall performance, there is a marked performance gap between R32B-0 and R32B, which further validates the effectiveness of \textit{financial domain knowledge expansion} and \textit{structured, logical reasoning distillation} in FEVO.

Our main contributions are summarized as follows:
\begin{itemize}
    \item We propose \textbf{FEVO}, a multi-stage enhancement framework for the financial domain. The pipeline consists of continued pretraining (CPT) to \textit{expand specialized domain knowledge}, supervised fine-tuning (SFT) to \textit{instill structured, elaborate reasoning}, and reinforcement learning (RL), which \textit{further integrates the financial domain knowledge acquired during CPT with the chain-of-thought reasoning patterns learned in SFT}, thereby pushing the upper limit of the model's reasoning capabilities in the financial domain.
    
    \item We propose a multi-source integration strategy (encompassing CPA textbooks, industry corpora, and mock exam questions) along with finely-grained data filtering, which includes answer-reference matching, reasoning chain validation, image/table filtering, and sub-question separation. To address the issue of ``reward hacking'' during RL, we also leverage DeepSeek-R1 to convert to convert single-/multiple-choice questions into open-ended ones and vastly decrease the probability of models ``guessing'' a correct answer. These efforts result in \textbf{FEVO-Train-C, Train-S,} and \textbf{Train-R}\textemdash high-quality datasets tailored to each stage's training objectives.
    
    \item We demonstrate that \textbf{FEVO-R32B}, built upon Qwen2.5-32B \citep{qwen25}, outperforms larger models such as GPT-4o and DeepSeek-R1 as well as specialized financial models like Dianjin-R1 on multiple financial benchmarks including FinanceIQ, CFLUE, and FIN-EVA. It also surpasses FEVO-R32B-0 (trained from Qwen2.5-32B-Instruct using only RL), validating the effectiveness of the FEVO framework in \textit{expanding financial domain knowledge} and \textit{establishing structured reasoning} to enable high-quality financial reasoning on medium-scale open-source LLMs.
\end{itemize}

\section{Dataset Curation}

\subsection{Data Source Selection}

Our data sources consist of open-source and in-house datasets dedicated to financial domain knowledge, complemented by general-purpose datasets that cover a wide spectrum of domains.

\begin{itemize}
    \item \textbf{Acc Learn} is an in-house dataset consisting of teaching material for registered accountants and CPA (Certified Public Accountants) examination preparations.
    \item \textbf{FinCorpus} \citep{FinCorpus} is an open-source financial dataset curated by Duxiaoman which contains financial news reports, listed company announcements, and financial exam questions. We utilize the latter portion of the dataset in our pipeline.
    \item \textbf{IndustryCorpus} \citep{IndustryCorpus} is an open-source pretraining dataset curated by BAAI that spans over 18 industries including medical, education, finance, law, etc. We utilize the finance category in our pipeline.
    \item \textbf{InfiMM-WebMath-Edu-zh} \citep{infimmwebmath40b2024} is an open-source Chinese mathematics pretraining dataset filtered from Bytedance's webMath 40B and curated by Finemath. It includes 2.4M webapges, 8.5M relevant pictures and 40B text tokens all filtered from Common-Crawl.
    \item \textbf{Dolma} \citep{soldaini2024dolmaopencorpustrillion} is a dataset of 3 trillion tokens from a diverse mix of web content, academic publications, code, books, and encyclopedic materials.
    \item \textbf{SkyPile 150B} \citep{skypile-150B} is a large dataset specifically designed for pre-training Chinese language models, with an overall token count of around 150B. It contains information extracted from various publicly-available Chinese websites which then undergoes a stringent process of data filtering, deduplication, and desensitization.
    \item \textbf{CFLUE} \citep{zhu-CFLUE-2024-benchmarking} (Chinese Financial Language Understanding Evaluation) is an open-source Chinese benchmark designed for the purpose of evaluating LLM performance across a wide spectrum of financial tasks. The \textit{Knowledge Assessment} component consists of more than 38,000 multiple-choice questions selected from 15 different types of financial qualification simulation exams along with detailed explanations; the \textit{Application Assessment} component provides over 16,000 instances covering five classic NLP tasks including text classification, machine translation, relation extraction, reading comprehension, and text generation.
    \item \textbf{Chinese-DeepSeek-R1-Distill-data-110k} \citep{Chinese-Data-Distill-From-R1} is a dataset curated for the purposes of distilation from DeepSeek-R1. We only use data from the applied mathematics portion with quality scores of 10.
\end{itemize}

\subsection{Data Processing and Synthesis}

To obtain high-quality training data, we first filter the gathered datasets. Our training pipeline consists of three stages: continued pretraining (CPT), supervised fine-tuning (SFT), and reinforcement learning (RL). We seek to achieve different goals with these stages, and therefore construct the corresponding datasets in different ways.

\subsubsection{CPT Data}

\begin{figure}[htbp]
\begin{center}
\includegraphics[width=0.8\textwidth]{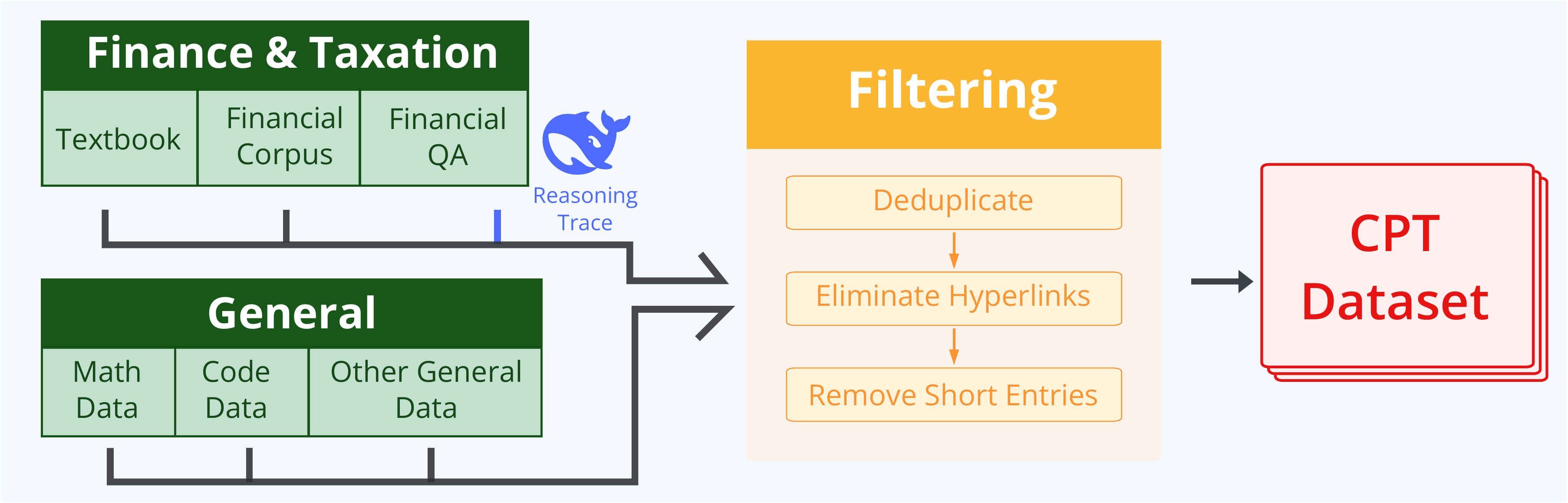}
\end{center}
\caption{Curation process for the CPT dataset}
\label{fig:cpt-data}
\end{figure}

In the CPT stage, our objective is to expand the trained model's knowledge \citep{ke2023continualpretraininglanguagemodels} in the financial domain, thus ensuring that it can recall sufficient amounts of relevent knowledge when solving financial tasks. Thus we utilize open-source financial text corpus to obtain copius amounts of domain data. 

One major weakness of pretrained models is insufficient keyword expansion capability for financial terms. When attempting to solve financial tasks, this prevents the model from accurately establishing connections to professional concepts that oftentimes contain complex logic, thereby undermining its performance. To address this issue, we leverage outside knowledge of frontier models along with multiple-choice questions within FinCorpus through the following process (shown in Figure \ref{fig:cpt-data}): for a given sample question, we ask DeepSeek-R1 (other frontier reasoning models also work) to convert the reference answer choice into a detailed reasoning response which thoroughly explains steps taken to arrive at the solution. After completing data filtering operations including deduplication, hyperlink removal, and short entry elimination, we append detailed reasoning trajectories to the CPT dataset. This allows us to \textit{leverage the external knowledge of frontier models to optimize reasoning patterns} and \textit{enhance the keyword expansion capabilities of trained models}.

We also include general-domain data in our CPT dataset for the following reasons:

\begin{itemize}
    \item \textbf{Stable output:} When compared to exclusively using financial domain data, integrating general domain data leads to more stable, coherent model responses after training.
    \item \textbf{Performance preservation:} Exclusive exposure to financial domain data during CPT leads to degraded performance at general tasks such as conversation. To prevent the trained model from devolving to a rigid answering bot for financial test questions, general domain data is used to preserve performance at out-of-domain tasks
\end{itemize}

After the above process, we obtain our CPT dataset \textbf{FEVO-Train-C}, with roughly 188M tokens.

\subsubsection{SFT Data}

\begin{figure}[htb]
\begin{center}
\includegraphics[width=\textwidth]{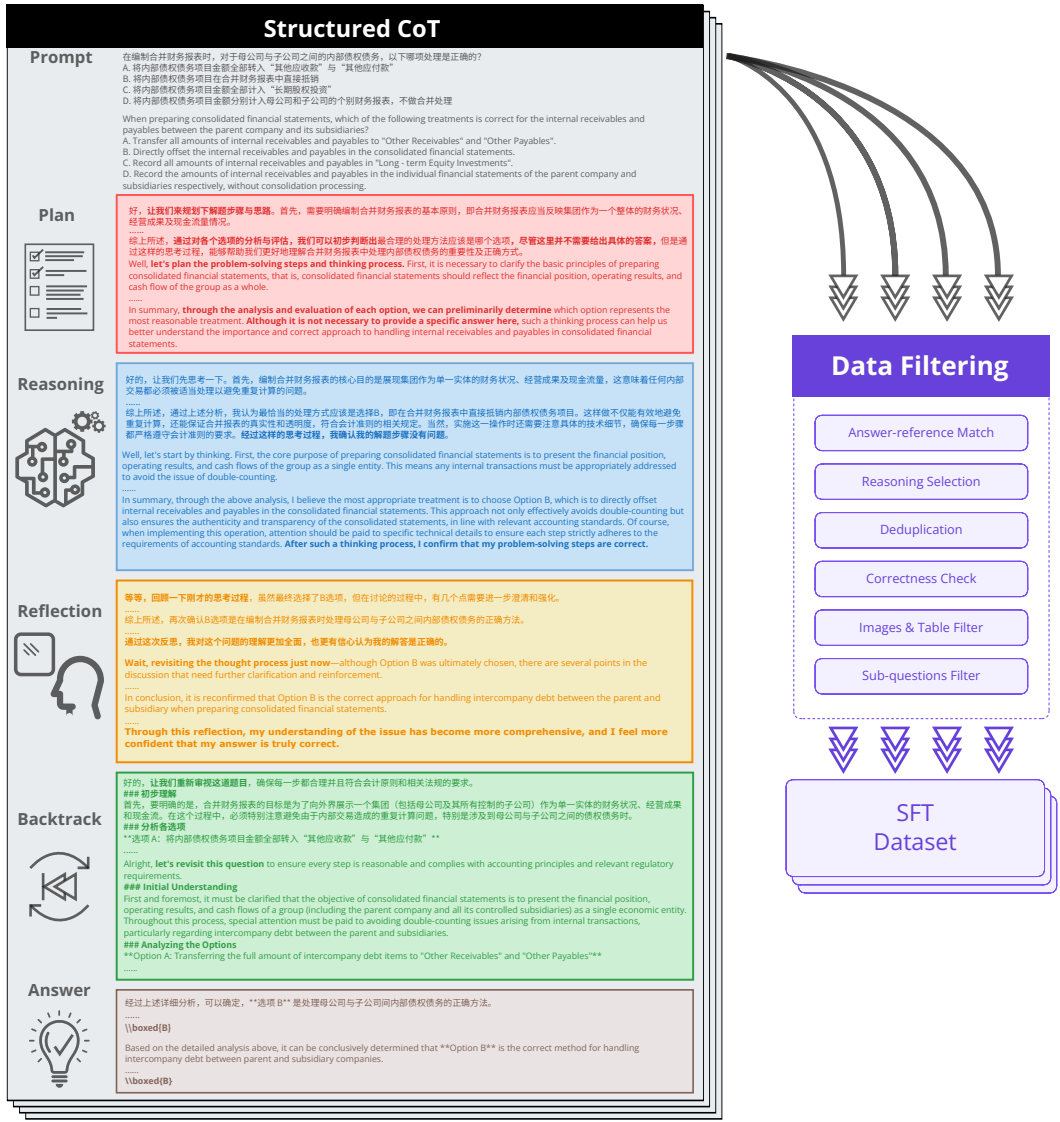}
\end{center}
\caption{Example of a structured CoT response and the filtering process to obtain SFT dataset}
\label{fig:sft-data}
\end{figure}

In the SFT stage, the goal is to guide the trained model toward utilizing an elaborate, intricate reasoning paradigm when answering questions. We utilized frontier reasoning models and prompt engineering to generate long, structured Chain-Of-Though (CoT) traces for knowledge distillation. The generated CoT traces are structured to contain the following elements:

\begin{itemize}
    \item \textbf{Plan:} An initial plan drafted to tackle the problem
    
    \item \textbf{Reasoning:} A comprehensive thinking process to solve the problem step by step.
    
    \item \textbf{Reflection:} A review of the reasoning steps in the previous stage, checking for potential errors and potentially overturning existing thought process
    
    \item \textbf{Backtracking:} An overall evaluation of all of the previous stages, utilizing potentially different approaches for solving attempt
    
    \item \textbf{Answer:} The final answer based upon all previous thought process
\end{itemize}

We obtain structured CoTs using two approaches: directly prompting DeepSeek-R1 to generate the CoT traces, and using Qwen2.5-72B-Instruct to generate the individual components in a multi-round session, then concatenate them to create the CoT traces. The data is then subjected through further filtering to eliminate entries not suitable for training (shown in Figure \ref{fig:sft-data}):

\begin{itemize}
    \item \textbf{Answer-Reference match:} Entries are removed where the generated answer does not match the provided reference answer
    \item \textbf{Reasoning Selection:} We check whether a) the answer matches the problem and b) the answer is a fluke response without a logical reasoning path, using GPT-4o as a judge. The entry is removed if either condition fails.
    \item \textbf{Deduplication:} We remove duplicate entries from the dataset using a combination of LSM and MinHash (N-gram is set to 13).
    \item \textbf{Correctness Check:} Due to noise introduced by response generation, it's possible that newly-formed response is incorrect. Thus, we verify that the prompt has a solution.
    \item \textbf{Images \& Table Filter:} Entries where the prompt contain image references, tables, and hyperlinks are removed.
    \item \textbf{Sub-questions Filter:} Entries that contain multiple sub-questions within the same prompt are removed.
\end{itemize}

The resulting dataset is \textbf{FEVO-Train-S}, comprised of 291K entries.

\subsubsection{RL Data}

\begin{figure}[htbp]
\begin{center}
\includegraphics[width=0.9\textwidth]{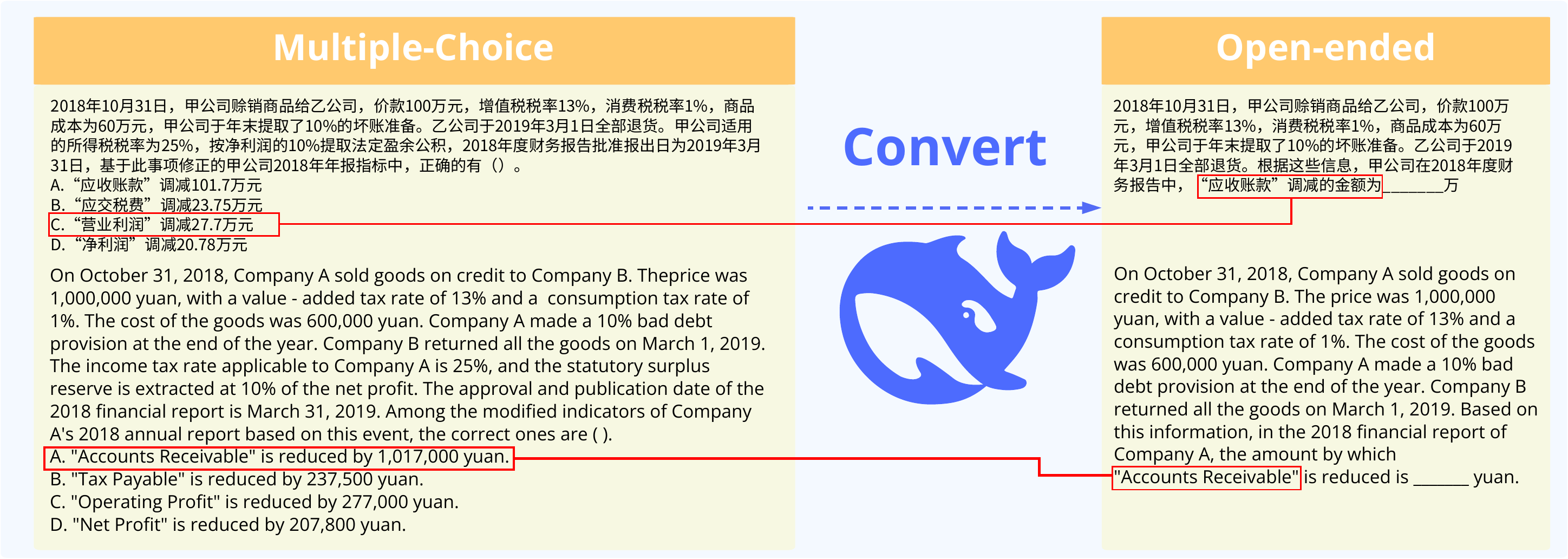}
\end{center}
\caption{Example of converting a single-choice question to an open-ended one. Necessary information is preserved from the original question, with one of the answer choices used in the new question}
\label{fig:prompt-convert}
\end{figure}

For the RL stage, we seek to further integrate the trained model's expanded knowledge (acquired during CPT) with its structured, CoT reasoning (acquired during SFT). We therefore utilize outcome rewards: a correct final response validates the entire reasoning trajectory, thus encouraging diverse reasoning pathway exploration for LLMs where more domain knowledge may be utilized. As such, the training dataset only need to consist of question-answer pairs.

A problem with our initial data is that many are single-/multiple-choice questions. This presents the risk of reward-hacking (\citealp{gao2022scalinglawsrewardmodel}; \citealp{everitt2021rewardtamperingproblemssolutions}), since the very limited sample space of potential solutions means that there is a considerable possibility for LLMs to guess the correct answer without actually understanding how to solve it. To address reward-hacking, we construct a pipeline for transforming fixed single-/multiple-choice questions into open-ended response questions (an example is shown in Figure \ref{fig:prompt-convert}). In addition to eliminating reward-hacking, this approach also enabled us to expand the training dataset by synthesizing multiple open-ended questions based on the different choices of the original question.

After obtaining the synthesized data, we further perform a combination of rule-based and model-based filtering to preserve only high-quality data for RL training. The filtering process is based on the following criteria:

\begin{itemize}
    \item \textbf{Clarity:} The question prompt should clearly convey a single request that leaves no room for further interpretations. 
    
    \item \textbf{Verifiability:} The reference answer should be verifiable and easily compared to a generated solution. 
    
    \item \textbf{Solvability:} The question prompt should be structured so that it has a correct solution. Due to the uneven quality of the raw data as well as potential noise introduced through model generation, the synthetic question prompt might not contain a valid solution.
    
    \item \textbf{Completeness:} The question prompt should contain all contextual information necessary to arrive at a solution. Note that contextual information in this case refers to the specific setup of the problem. \textit{We do not consider a prompt to be lacking information if it asks for well-known domain knowledge (e.g. content of relevant laws or standard industry practice).}
\end{itemize}

After these steps, we obtain RL dataset \textbf{FEVO-Train-R}, with 32k entries of clear, solvable, open-ended questions.

\section{Training Pipeline}

\subsection{Knowledge Expansion via CPT}

Many problems in the financial domain are highly specialized, and solving them requires significant amounts of relevant background knowledge. Current open-source base models like Qwen \citep{qwen3} and LLama \citep{llama4} are trained to handle general tasks, meaning that they have been exposed to data from an incredibly wide spectrum of domains throughout their training process. This reduces their ability to correctly recall sufficient information when answering questions from the financial domain.

To address this lack of innate knowledge, we utilize continued pre-training (CPT) to expand the relevant knowledge of LLMs. Using substantial financial data ranging from examination questions to real-world news reports, the trained LLM gains deeper understanding of financial domain knowledge and establishes more intricate connections between relevant financial information.

\subsection{Structured Reasoning Distillation via SFT}

After CPT, the model's financial domain knowledge is greatly expanded, but the response it generates is often inconsistent. We therefore utilize supervised fine-tuning (SFT) and our SFT reasoning datasets \textbf{FEVO-Train-S} to guide the model towards generating coherent responses with intelligible reasoning traces, using high-quality responses generated by frontier models to distill knowledge into the model \citep{hinton2015knowledgedistill}. 

During the fine-tuning stage, for a sample question the model is instructed to generate both a reasoning trace and a final output. This complete output is then compared against the reference answer (which contains an elaborate and intricate reasoning trace generated by a frontier model) to calculate a loss that is used for parameter updates. Studies have shown that this act of knowledge distillation can effectively promote reasoning patterns and significant improve performance on reasoning tasks \citep{ye2025limoreasoning,muennighoff2025s1simpletesttimescaling}.

During preliminary experiments, we observe that fine-tuned models at this stage already exhibit well-structured, logic CoTs when responding to questions. 

\subsection{Knowledge-Reasoning Integration via RL}

For RL training, we utilize a modified version of VAPO \citep{yue2025vapoefficientreliablereinforcement}\textemdash Value-Augmented Proximal Policy Optimization\textemdash with sampling enhancements, as well as a rule-based reward mechanism with model complements. We implement our modifications in the veRL framework \citep{sheng2024hybridflow-verl}.

The standard VAPO algorithm (\ref{vapo-explain}) does not employ any sampling and rollout techniques, which undermines effective use of training data and potentially leads to slower training with worse results. As such, we introduce \textbf{Balanced Batching}.

\subsubsection{Balanced Batching}

Value-based RL algorithms like PPO \citep{schulman2017ppo} allow for policy model updates even for sampled questions that elicit all-correct or all-wrong responses from the trained policy. But the resulting sparse reward signals from these responses make it difficult for the trained policy to learn effectively. Therefore, we eliminate these kind of questions \citep{yu2025dapoopensourcellmreinforcement} from the training batch to ensure that an informative batch is utilized for model parameter update.

However, filtering all-correct and all-wrong questions means that many generated responses are discarded, thereby necessitating multiple rollouts to acquire sufficient samples for the model update training batch. This issue becomes more pronounced as the model becomes more capable throughout training, with our preliminary experiments finding that percentage of all-correct responses in a single generation batch steadily increases to as high as 70\%, leading to significant time expenditure as rollout is repeated more often. To address this issue, we construct the rollout batch based on existing accuracy records to include questions that are most likely to be utilized for model updates. We explored two approaches to the filtering condition:

\begin{itemize}
    \item Only keep questions with an existing accuracy that falls in between accuracy thesholds $m_{hard}$ and $m_{easy}$ in the rollout batch
    \item Define accuracy thresholds $m_{hard}, \ m_{easy}$ for hard and easy questions, and impose upper limits $L_{hard}, \ L_{easy}$ (defined as a percentage of the rollout batch size) on their frequency in the rollout batch
\end{itemize}

After experimenting with the two approaches, we utilized the latter in our RL training process due to greater flexibility, thus obtaining \textit{balanced batches for rollout}. During RL training, we observed an average training speed improvement of 20\% compared to default dynamic sampling, with this improvement becoming even more pronounced after several epochs into the training process. We also found that balanced batching have minimal effects on the performance evolution of the model during our experiments, indicating that it doesn't filter out much data that could potentially be utilized for model updates.

\subsubsection{Reward mechanism}

For the reward aspect of RL, our reward mechanism incorporates a merged reward for accuracy and format, as well as a language-consistency reward.

\begin{itemize}
    \item \textbf{Accuracy and Formatting reward:} We construct the format reward to guide the model towards generating well-structured correct responses. In the system prompt, we instruct the model to respond by generating a thinking process enclosed between \texttt{<think>} and \texttt{</think>} tags, then placing the final answer within \texttt{\textbackslash boxed\{\}}. If the generated response is not in this format, then a penalty of \texttt{-2.0} is applied and the correctness of the response will not be checked. If the format is deemed to be correct, then the generated response within \texttt{\textbackslash boxed\{\}} will be evaluated for accuracy. If it's equivalent to the reference answer, a reward of \texttt{2.0} is given; otherwise, a penalty of \texttt{-1.5} is given.

    However, in the scenario where both the generated and reference answers are long, written rules might not catch all edge cases. Therefore, we also utilize model reasoning (by calling a LLM API) as an additional judge when the long response is deemed false by set rules. 
    $$R_{format + accuracy} = \begin{cases}
        -2.0 & \text{if format incorrect} \\
        -1.5 & \text{if format correct, answer incorrect} \\
        2.0 & \text{if format and answer both correct}
    \end{cases}$$
    
    \item \textbf{Language consistency reward:} We define a language consistency reward to ensure that the model responds consistently in either Chinese or English. Specifically, for a given model response, the percentage of Chinese tokens $p_{cn}$ and English tokens $p_{en}$ is calculated. If both $p_{cn}$ and $p_{en}$ are below a predetermined threshold $k$, we apply a penalty of \texttt{-0.5}.

    $$R_{language} = \begin{cases}
        -0.5 & \text{if} \quad p_{cn} < k \ \cap \ p_{en} < k\\
        0.0 & \text{if} \quad p_{cn} \geq k \ \cup p_{en} \geq k
    \end{cases}$$
\end{itemize}

\section{Experiments}

\subsection{Setup}

The base model used for our FEVO training framework is Qwen2.5-32B, a non-reasoning model that demonstrates excellent performance at following system prompt instructions. Utilizing the curated datasets, we conduct training using the following parameters:

\subsubsection{Training parameters}

During the CPT stage, we used a learning rate of $3\times 10^{-4}$, a warm-up of 0.02, a train batch size of 4, and a split size of 2048. The model is then trained for 1 epoch, resulting to \textbf{FEVO-C32B}.

During the SFT process, the following parameters were utilized: A learning rate of $5 \times 10^{-5}$, a warm-up of 0.02, and a train batch size of 1. We apply a cutoff length of 8192 for generated responses, and then trained the model for 1 epoch to obtain \textbf{FEVO-S32B}.

For the RL stage, we chose a value model checkpoint from preliminary RL experiments on the same training set. This value model has a tracked loss less than 1 and an explained variance greater than 0.5. We then allowed this value model to warm up for 20 steps before policy model training to ensure that it is accurate enough at the beginning of policy model training, since this period is crucial to determining subsequent policy model evolution behavior. For de-coupled and length-adaptive GAE, we set $\lambda_{critic} = 1$ for value model updates, and $\lambda_{actor} = 1 - \frac{1}{\alpha l}$ where $\alpha=0.16$ to update the policy model. The clip range is set to $\epsilon_{lower}=0.2$ and $\epsilon_{higher}=0.28$. We used 128 prompts per train batch, generated 16 responses for each prompt, and set mini-batch size to 128. For balanced batching, we defined thesholds $m_{hard} = 0.0, m_{easy} = 0.95$ for hard and easy questions, then set corresponding limits as $L_{hard} = 0.1 \cdot \textit{rollout batch size}, L_{easy} = 0.3 \cdot \textit{rollout batch size}$. We set the language consistency threshold $k = 0.8$. We maintain training until rewards stagnate, then chose the best-performing checkpoint as \textbf{FEVO-R32B}.

\subsubsection{Evaluation}

\begin{table}[htb]
    \centering
    \scriptsize
    \begin{tabular}{
    >{\RaggedRight\arraybackslash}p{22ex} 
    >{\Centering\arraybackslash}p{8ex}
    >{\Centering\arraybackslash}p{8ex}
    >{\Centering\arraybackslash}p{6ex}
    >{\Centering\arraybackslash}p{8ex}
    >{\Centering\arraybackslash}p{6ex}
    >{\Centering\arraybackslash}p{12ex}
    >{\Centering\arraybackslash}p{10ex}
    >{\Centering\arraybackslash}p{6ex}}
        \toprule
        \multirow{2}{*}{Model} & \multicolumn{6}{c}{Finance} & {General} & \multirow{2}{*}{Avg.} \\
        \cmidrule(lr){2-7} \cmidrule(lr){8-8}
        & {Fin CPA} & {Fin CCR} & {Fin IQ} & {CFLUE} & {FIN-EVA} & {OpenFinData} & {MATH 500} & \\
        \midrule
        \rowcolor{mygray} \multicolumn{9}{c}{Reasoning models} \\
        \midrule        
        QwQ                           & 57.6 & 84.72 & 74.53 & 83.49 & 61.90 & 80.00 & 91.00 & 76.18 \\
        Qwen3-32B-MOE (w/ think)      & 52.0 & 84.72 & 77.67 & 76.37 & 75.20 & 80.15 & \underline{92.00} & 76.87 \\
        DeepSeek-R1-Distill-Qwen-32B  & 50.4 &  77.74 & 77.09 & 76.86 & 76.39 & 77.69 & 89 & 75.02 \\
        DeepSeek-R1                   & 55.2 & 79.0 & 78.23 & 83.49 & 73.00 & 80.15 & \textbf{94.80} & 77.70 \\
        \midrule
        \rowcolor{mygray} \multicolumn{9}{c}{Non-reasoning models} \\
        \midrule
        Qwen2.5-32B                 & 20.0 & 50.50 & 78.23 & 60.23 & 68.00 & 67.00 & 69.00 & 59.00 \\
        Qwen2.5-32B-Instruct        & 51.2 & 77.24 & 77.67 & 77.95 & 74.70 & 80.00 & 81.00 & 74.25 \\
        Qwen3-32B-MOE (w/out think) & 55.2 & 67.61 & 75.10 & 78.20 & 76.00 & 79.23 & 89.10 & 74.35 \\
        Qwen2.5-72B-Instruct        & 50.0 & 79.57 & \underline{80.24} & 79.46 & 76.00 & 79.53 & 82.20 & 75.29 \\
        GPT-4o                      & 45.6 & 54.00 & 71.45 & 71.68 & 50.00 & 78.92 & 75.90 & 63.94 \\
        \midrule
        \rowcolor{mygray} \multicolumn{9}{c}{Specialist models} \\
        \midrule
        Llama3-XuanYuan3-70B-Chat   & 26.0 & 58.14 & 74.02 & 71.25 & 70.90 & 66.77 & 62.10 & 61.31 \\
        XuanYuan-FinX1-Preview      & 49.0 & 73.09 & 80.01 & 78.10 & 64.10 & 78.92 & 80.00 & 71.89 \\
        Fin-R1 (7B)                 & 26.0 & 59.47 & 66.92 & 68.45 & 67.90 & 76.17 & 70.20 & 62.16 \\
        Dianjin-R1-32B                 & 60.8 & \underline{85.38} & 62.40 & \textbf{86.75} & 76.00 & \underline{85.85} & 88.20 & \underline{77.91} \\
        \midrule
        \rowcolor{mygray} \multicolumn{9}{c}{FEVO models} \\
        \midrule
        FEVO-R32B-0     & \underline{63.04} & 78.70 & 78.17 & 82.58 & \underline{85.1} & 82.45 & 74.76 & 77.83 \\
        FEVO-C32B       & 57.0 & 66.40 & 68.9 & 67.0 & 59.0 & 79.23 & 76.2 & 67.68 \\
        FEVO-S32B       & 58.4 & 78.24 & 77.69 & 78.1 & 82.9 & 77.24 & 84.00 & 76.65 \\
        FEVO-R32B       & \textbf{73.60} & \textbf{88.20} & \textbf{86.14} & \underline{85.2} & \textbf{85.66} & \textbf{87.70} & 88.6 & \textbf{85.01} \\
        \bottomrule
    \end{tabular}

\caption{\textit{avg@10} (unless otherwise stated) performance for models across benchmarks. Performance is reported as Accuracy (\%), best results are in \textbf{bold}, second best in \underline{underline}.}
\label{tab:benchmark_results_main}
\end{table}

We select benchmarks from both the financial domain as well as other domains to evaluate the performance of our model. For financial tests, we used Finance IQ \citep{FinanceIQ}, CFLUE \citep{zhu-CFLUE-2024-benchmarking}, FIN-EVA \citep{FIN-EVA}, OpenFinData \citep{openfindata}, and our in-house Fin CPA. We also curated \textbf{Fin CCR} (Chinese Computation \& Reasoning), a test dataset of 602 entries compiled exclusively of computation and reasoning-focused financial questions obtained from FIN-EVA and OpenFinData. To test performance on out-of-domain tasks, we used MATH 500 \citep{hendrycksmath2021}.

We evaluate our CPT model (FEVO-C32B), SFT model (FEVO-S32B), and final RL model (FEVO-R32B) on these datasets. To determine the effectiveness of our CPT and SFT stages, we also train Qwen2.5-32B-Instruct on our dataset using solely RL to obtain FEVO-R32B-0 and evaluate its performance.

For comparison purposes, we include evaluation results of leading reasoning models (like DeepSeek-R1\footnote{Version used is Jan 20th 2025} and QwQ) and specialist models like DianJin-R1-32B and Fin-R1. 

Where possible, models are evaluated using vLLM version 0.8.2 with a batch size of 10,000 and \texttt{n=1} for number of responses. The generation temperature is set to 0.6 for reasoning models, and 0.0 for non-reasoning models. 

\subsection{Results}
Table \ref{tab:benchmark_results_main} presents the \textit{(avg@10)} evaluation results for both financial and general domain tasks.

For financial domain tasks, we observe consistently improving performance for our models at different stages of the training pipeline, with \textbf{FEVO-R32B} reaching top performance in most benchmarks and beating much larger models like GPT-4o and DeepSeek-R1 . It also shows better performance than the specialist models as well as FEVO-R32B-0 (trained from Qwen2.5-32B-Instruct), validating the effectiveness of using CPT to expand the base model's domain knowledge.

\section{Discussions}

\textbf{Structured reasoning contributes greatly to increased accuracy}. During RL training, we observed progressively increasing response lengths for both FEVO-R32B-0 and FEVO-R32B, signifying the emergence of more complex and elaborate reasoning during problem-solving attempts. However, although these two models ended up generating responses of similar length to most questions, there exists a noticeable gap between their benchmark performance. Knowledge gap in the financial domain contributes to this, but the composition of their reponses (shown in \ref{sample-comparison}) also played a big role. As shown in the figure, FEVO-R32B-0's response expanded upon all answer choices in detail before summarizing and deciding upon a final solution. FEVO-R32B on the other hand generated a lot more reflections and self-questioning in its attempt to answer the question. The constant reflection and back-tracking allowed FEVO-R32B to analyze questions more holistically and expand upon relevant knowledge, thereby resulting in higher-quality results.

\textbf{Difficulty of training data significantly influences response length}. During preliminary RL experiments using Qwen2.5-32B-Instruct and vanilla VAPO, we observed that although test performance was steadily increase, the response length did not increase along with it. Rather, the response length exhibited very little change throughout training, only oscillating slightly. It was only after we applied dynamic sampling \citep{yu2025dapoopensourcellmreinforcement} and discarded all-correct/all-wrong responses in the training batch that we began to observe meaningful increase in response length. Diving deeper into the discarded and retained samples, we found that responses to more difficult questions (lower accuracy) are consistently longer than those for simpler ones. By using dynamica sampling to discard short responses to trivial questions, the model is gradually guided towards generating longer responses throughout the training process.

\textbf{Too much difficult questions for training lead to degraded performance}. On the subject of dynamic sampling, it can be viewed as a form a curriulum learning (structures training by sampling questions in an organized progression, typically from easy to difficult), which has been used to great effects in many works \citep{kimiteam2025kimik15}. However, we found that applying this during the RL phase actually caused the performance of our model to degrade. In our RL setting, a larger upper limit $L_{hard}$ for difficult questions in the training batch produced worse training trajectory than a smaller limit. This suggests that a balanced composition of questions is more useful then strictly difficult questions that models can barely answer correctly.

\textbf{Generation temperature has major effect on the output of reasoning models}. Another interesting phenomenon we observed was that setting \textit{temperature=0} for reasoning models (both FEVO and other models) frequently resulted in responses that keep on repeating a certain portion of itself and failing to answer the actual question. When we increase the temperature to common values like 0.6, this phenomenon is largely suppressed. For short CoT models, setting temperature to 0 didn't elicit the above phenomenon. After further investigations, we found that this issue is because reasoning models have fairly high likelihood of generating key words such as \textbf{wait}, \textbf{however}, and \textbf{maybe}. These key words signifies intermediate reasoning processes during problem-solving which enables reasoning models to produce more elaborate responses. Since a temperature of 0 would make a model choose the most likely token, reasoning models would therefore have a high probability of generating these reasoning tokens, thus getting stuck in reasoning loops and failing to answer the question.

\section{Related Works}

We review existing works that deal with training LLMs specifically for the financial domain.

\textbf{XuanYuan-FinX1-Preview} \citep{xuanyuan-finx1} is trained on top of XuanYuan3.0. It utilizes a two-stage process of SFT and RL. For the RL stage, the PPO algorithm was used in conjunction with a dedicated process reward model (PRM) and outcome reward model (ORM).

Qian et al. \citep{qian2025fino1transferabilityreasoningenhanced} address the critical challenges of scarce high-fidelity CoT datasets and missing evaluation benchmarks in financial reasoning by constructing the financial-specific CoT corpus FinCoT and comprehensive benchmark FinReason. They further develop the \textbf{Fin-o1} model through a framework integrating supervised fine-tuning with GRPO-based reinforcement learning.

\textbf{Fin-R1} \citep{fin-r1} and \textbf{Dianjin-R1} \citep{dianjin-r1} are recent works that produced specialist models focused in the financial domain. These two works both trained their models through a combination of SFT and RL with GRPO, using consistent rule-based approach to construct SFT and RL datasets that leverages both English as well as Chinese data. Their RL algorithm GRPO does not utilize a value model in estimating trajectory advantages \citep{shao2024deepseekmathpushinglimitsmathematical}; instead the algorithm relies on generating multiple responses to the sampled question, then utilizing the mean and standard deviation of the rewards to compute estimated advantages. Through their training frameworks, they were able to produce relatively small models (7B and 32B parameter size) that achieved competitive performance on financial benchmarks, even when measured against much larger models like GPT-4o and DeepSeek-R1.

\section{Conclusion}

We introduce \textbf{FEVO}, a multi-stage post-training framework dedicated to enhancing LLM performance on financial and taxation tasks. It incorporates CPT to \textit{expand the domain knowledge of LLMs}, SFT to \textit{establish structured, elaborate reasoning}, and RL to \textit{further integrate the newly-obtained knowledge with learned reasoning structure in order to push the performance boundaries of LLMs}. This pipeline effectively improves LLM performance on intricate and highly specialized financial questions, as demonstrated by our trained models' performance on curated financial benchmarks, where \textbf{FEVO-R32B} obtained state-of-the-art performance on several benchmarks, out-performing much larger LLMs and specialist models in the process.

For future work, we plan to conduct more comprehensive research in three directions. First, developing better RL optimization algorithms that are more efficient and effective will significantly reduce the computational and time constraints currently limiting RL training. Second, we plan to explore more challenging tasks in the financial domain (e.g., questions that require generating elaborate graphs or legally-compliant documents) to push the boundaries for what models can accomplish. And finally, developing better reward models that can provide more finely-grained rewards than current rule-based implementations would better guide the self-evolution process of models during RL training.

\bibliography{iclr2025_conference}

\begin{thebibliography}{36}
\providecommand{\natexlab}[1]{#1}
\providecommand{\url}[1]{\texttt{#1}}
\expandafter\ifx\csname urlstyle\endcsname\relax
  \providecommand{\doi}[1]{doi: #1}\else
  \providecommand{\doi}{doi: \begingroup \urlstyle{rm}\Url}\fi

\bibitem[AI(2025)]{llama4}
Meta AI.
\newblock The llama 4 herd: The beginning of a new era of natively multimodal ai innovation.
\newblock \url{https://ai.meta.com/blog/llama-4-multimodal-intelligence/}, 2025.

\bibitem[BAAI(2024)]{IndustryCorpus}
BAAI.
\newblock Fincorpus.
\newblock \url{https://huggingface.co/datasets/BAAI/IndustryCorpus}, 2024.

\bibitem[DeepSeek-AI(2025)]{deepseek-r1}
DeepSeek-AI.
\newblock Deepseek-r1: Incentivizing reasoning capability in llms via reinforcement learning, 2025.
\newblock URL \url{https://arxiv.org/abs/2501.12948}.

\bibitem[Everitt et~al.(2021)Everitt, Hutter, Kumar, and Krakovna]{everitt2021rewardtamperingproblemssolutions}
Tom Everitt, Marcus Hutter, Ramana Kumar, and Victoria Krakovna.
\newblock Reward tampering problems and solutions in reinforcement learning: A causal influence diagram perspective, 2021.
\newblock URL \url{https://arxiv.org/abs/1908.04734}.

\bibitem[Gao et~al.(2022)Gao, Schulman, and Hilton]{gao2022scalinglawsrewardmodel}
Leo Gao, John Schulman, and Jacob Hilton.
\newblock Scaling laws for reward model overoptimization, 2022.
\newblock URL \url{https://arxiv.org/abs/2210.10760}.

\bibitem[Group(2023)]{FIN-EVA}
Ant Group.
\newblock Fin-eva version 1.0.
\newblock \url{https://github.com/alipay/financial_evaluation_dataset}, 2023.

\bibitem[Han et~al.(2024)Han, Jian, Hu, Liu, Wang, Fan, Ai, Huang, He, Yang, and You]{infimmwebmath40b2024}
Xiaotian Han, Yiren Jian, Xuefeng Hu, Haogeng Liu, Yiqi Wang, Qihang Fan, Yuang Ai, Huaibo Huang, Ran He, Zhenheng Yang, and Quanzeng You.
\newblock Infimm-webmath-40b: Advancing multimodal pre-training for enhanced mathematical reasoning, 2024.
\newblock URL \url{https://arxiv.org/abs/2409.12568}.

\bibitem[Hendrycks et~al.(2021)Hendrycks, Burns, Kadavath, Arora, Basart, Tang, Song, and Steinhardt]{hendrycksmath2021}
Dan Hendrycks, Collin Burns, Saurav Kadavath, Akul Arora, Steven Basart, Eric Tang, Dawn Song, and Jacob Steinhardt.
\newblock Measuring mathematical problem solving with the math dataset.
\newblock \emph{NeurIPS}, 2021.

\bibitem[Hinton et~al.(2015)Hinton, Vinyals, and Dean]{hinton2015knowledgedistill}
Geoffrey Hinton, Oriol Vinyals, and Jeff Dean.
\newblock Distilling the knowledge in a neural network, 2015.
\newblock URL \url{https://arxiv.org/abs/1503.02531}.

\bibitem[Hu et~al.(2025)Hu, Zhang, Han, Jiang, Zhang, and Shum]{hu2025openreasonerzeroopensourceapproach}
Jingcheng Hu, Yinmin Zhang, Qi~Han, Daxin Jiang, Xiangyu Zhang, and Heung-Yeung Shum.
\newblock Open-reasoner-zero: An open source approach to scaling up reinforcement learning on the base model, 2025.
\newblock URL \url{https://arxiv.org/abs/2503.24290}.

\bibitem[Ji et~al.(2025)Ji, Zhao, Tian, Wang, Chen, Peng, Zhao, and Li]{2025diff-aware-rl}
Yunjie Ji, Sitong Zhao, Xiaoyu Tian, Haotian Wang, Shuaiting Chen, Yiping Peng, Han Zhao, and Xiangang Li.
\newblock How difficulty-aware staged reinforcement learning enhances llms' reasoning capabilities: A preliminary experimental study, 2025.
\newblock URL \url{https://arxiv.org/abs/2504.00829}.

\bibitem[Ke et~al.(2023)Ke, Shao, Lin, Konishi, Kim, and Liu]{ke2023continualpretraininglanguagemodels}
Zixuan Ke, Yijia Shao, Haowei Lin, Tatsuya Konishi, Gyuhak Kim, and Bing Liu.
\newblock Continual pre-training of language models, 2023.
\newblock URL \url{https://arxiv.org/abs/2302.03241}.

\bibitem[Liu et~al.(2025{\natexlab{a}})Liu, Wang, Shen, Peng, Zhang, Du, and Wang]{Chinese-Data-Distill-From-R1}
Cong Liu, Zhong Wang, ShengYu Shen, Jialiang Peng, Xiaoli Zhang, ZhenDong Du, and YaFang Wang.
\newblock The chinese dataset distilled from deepseek-r1-671b.
\newblock \url{https://huggingface.co/datasets/Congliu/Chinese-DeepSeek-R1-Distill-data-110k}, 2025{\natexlab{a}}.

\bibitem[Liu et~al.(2025{\natexlab{b}})Liu, Guo, Lou, Zeng, Niu, Wang, Xu, Cai, Yang, Zhao, Li, Xu, Chen, Chen, Bai, and Zhang]{fin-r1}
Zhaowei Liu, Xin Guo, Fangqi Lou, Lingfeng Zeng, Jinyi Niu, Zixuan Wang, Jiajie Xu, Weige Cai, Ziwei Yang, Xueqian Zhao, Chao Li, Sheng Xu, Dezhi Chen, Yun Chen, Zuo Bai, and Liwen Zhang.
\newblock Fin-r1: A large language model for financial reasoning through reinforcement learning, 2025{\natexlab{b}}.
\newblock URL \url{https://arxiv.org/abs/2503.16252}.

\bibitem[Money \& Lab(2024)Money and Lab]{openfindata}
East Money and Shanghai~AI Lab.
\newblock Openfindata - llm open-source financial test dataset.
\newblock \url{https://github.com/open-compass/OpenFinData}, 2024.
\newblock Accessed: 2025-06-30.

\bibitem[Muennighoff et~al.(2025)Muennighoff, Yang, Shi, Li, Fei-Fei, Hajishirzi, Zettlemoyer, Liang, Candès, and Hashimoto]{muennighoff2025s1simpletesttimescaling}
Niklas Muennighoff, Zitong Yang, Weijia Shi, Xiang~Lisa Li, Li~Fei-Fei, Hannaneh Hajishirzi, Luke Zettlemoyer, Percy Liang, Emmanuel Candès, and Tatsunori Hashimoto.
\newblock s1: Simple test-time scaling, 2025.
\newblock URL \url{https://arxiv.org/abs/2501.19393}.

\bibitem[OpenAI(2024)]{openai-o1}
OpenAI.
\newblock Learning to reason with llms, 2024.
\newblock URL \url{https://openai.com/index/learning-to-reason-with-llms/}.
\newblock Last accessed 19 May 2025.

\bibitem[Qian et~al.(2025)Qian, Zhou, Wang, Peng, Yi, Huang, Xie, and Nie]{qian2025fino1transferabilityreasoningenhanced}
Lingfei Qian, Weipeng Zhou, Yan Wang, Xueqing Peng, Han Yi, Jimin Huang, Qianqian Xie, and Jianyun Nie.
\newblock Fino1: On the transferability of reasoning enhanced llms to finance, 2025.
\newblock URL \url{https://arxiv.org/abs/2502.08127}.

\bibitem[Qwen et~al.(2025)Qwen, Yang, Yang, Zhang, Hui, Zheng, Yu, Li, Liu, Huang, Wei, Lin, Yang, Tu, Zhang, Yang, Yang, Zhou, Lin, Dang, Lu, Bao, Yang, Yu, Li, Xue, Zhang, Zhu, Men, Lin, Li, Tang, Xia, Ren, Ren, Fan, Su, Zhang, Wan, Liu, Cui, Zhang, and Qiu]{qwen25}
Qwen, An~Yang, Baosong Yang, Beichen Zhang, Binyuan Hui, Bo~Zheng, Bowen Yu, Chengyuan Li, Dayiheng Liu, Fei Huang, Haoran Wei, Huan Lin, Jian Yang, Jianhong Tu, Jianwei Zhang, Jianxin Yang, Jiaxi Yang, Jingren Zhou, Junyang Lin, Kai Dang, Keming Lu, Keqin Bao, Kexin Yang, Le~Yu, Mei Li, Mingfeng Xue, Pei Zhang, Qin Zhu, Rui Men, Runji Lin, Tianhao Li, Tianyi Tang, Tingyu Xia, Xingzhang Ren, Xuancheng Ren, Yang Fan, Yang Su, Yichang Zhang, Yu~Wan, Yuqiong Liu, Zeyu Cui, Zhenru Zhang, and Zihan Qiu.
\newblock Qwen2.5 technical report, 2025.
\newblock URL \url{https://arxiv.org/abs/2412.15115}.

\bibitem[Schulman et~al.(2017)Schulman, Wolski, Dhariwal, Radford, and Klimov]{schulman2017ppo}
John Schulman, Filip Wolski, Prafulla Dhariwal, Alec Radford, and Oleg Klimov.
\newblock Proximal policy optimization algorithms, 2017.
\newblock URL \url{https://arxiv.org/abs/1707.06347}.

\bibitem[Schulman et~al.(2018)Schulman, Moritz, Levine, Jordan, and Abbeel]{schulman2018gae}
John Schulman, Philipp Moritz, Sergey Levine, Michael Jordan, and Pieter Abbeel.
\newblock High-dimensional continuous control using generalized advantage estimation, 2018.
\newblock URL \url{https://arxiv.org/abs/1506.02438}.

\bibitem[Shao et~al.(2024)Shao, Wang, Zhu, Xu, Song, Bi, Zhang, Zhang, Li, Wu, and Guo]{shao2024deepseekmathpushinglimitsmathematical}
Zhihong Shao, Peiyi Wang, Qihao Zhu, Runxin Xu, Junxiao Song, Xiao Bi, Haowei Zhang, Mingchuan Zhang, Y.~K. Li, Y.~Wu, and Daya Guo.
\newblock Deepseekmath: Pushing the limits of mathematical reasoning in open language models, 2024.
\newblock URL \url{https://arxiv.org/abs/2402.03300}.

\bibitem[Sheng et~al.(2024)Sheng, Zhang, Ye, Wu, Zhang, Zhang, Peng, Lin, and Wu]{sheng2024hybridflow-verl}
Guangming Sheng, Chi Zhang, Zilingfeng Ye, Xibin Wu, Wang Zhang, Ru~Zhang, Yanghua Peng, Haibin Lin, and Chuan Wu.
\newblock Hybridflow: A flexible and efficient rlhf framework.
\newblock \emph{arXiv preprint arXiv: 2409.19256}, 2024.

\bibitem[Soldaini et~al.(2024)Soldaini, Kinney, Bhagia, Schwenk, Atkinson, Authur, Bogin, Chandu, Dumas, Elazar, Hofmann, Jha, Kumar, Lucy, Lyu, Lambert, Magnusson, Morrison, Muennighoff, Naik, Nam, Peters, Ravichander, Richardson, Shen, Strubell, Subramani, Tafjord, Walsh, Zettlemoyer, Smith, Hajishirzi, Beltagy, Groeneveld, Dodge, and Lo]{soldaini2024dolmaopencorpustrillion}
Luca Soldaini, Rodney Kinney, Akshita Bhagia, Dustin Schwenk, David Atkinson, Russell Authur, Ben Bogin, Khyathi Chandu, Jennifer Dumas, Yanai Elazar, Valentin Hofmann, Ananya~Harsh Jha, Sachin Kumar, Li~Lucy, Xinxi Lyu, Nathan Lambert, Ian Magnusson, Jacob Morrison, Niklas Muennighoff, Aakanksha Naik, Crystal Nam, Matthew~E. Peters, Abhilasha Ravichander, Kyle Richardson, Zejiang Shen, Emma Strubell, Nishant Subramani, Oyvind Tafjord, Pete Walsh, Luke Zettlemoyer, Noah~A. Smith, Hannaneh Hajishirzi, Iz~Beltagy, Dirk Groeneveld, Jesse Dodge, and Kyle Lo.
\newblock Dolma: an open corpus of three trillion tokens for language model pretraining research, 2024.
\newblock URL \url{https://arxiv.org/abs/2402.00159}.

\bibitem[Team(2023{\natexlab{a}})]{FinCorpus}
Duxiaoman~DI Team.
\newblock Fincorpus.
\newblock \url{https://huggingface.co/datasets/Duxiaoman-DI/FinCorpus/tree/main/data}, 2023{\natexlab{a}}.

\bibitem[Team(2023{\natexlab{b}})]{FinanceIQ}
Duxiaoman~DI Team.
\newblock Financeiq.
\newblock \url{https://huggingface.co/datasets/Duxiaoman-DI/FinanceIQ}, 2023{\natexlab{b}}.

\bibitem[Team(2024)]{xuanyuan-finx1}
Duxiaoman~DI Team.
\newblock Xuanyuan-finx1-preview.
\newblock \url{https://github.com/ Duxiaoman-DI/XuanYuan}, 2024.
\newblock Accessed: 2024-03-18.

\bibitem[Team et~al.(2025)Team, Du, Gao, Xing, Jiang, Chen, Li, Xiao, Du, Liao, Tang, Wang, Zhang, Yuan, Lu, Tang, Sung, Wei, Lai, Guo, Zhu, Ding, Hu, Yang, Zhang, Yao, Zhao, Lu, Li, Yu, Gao, Zheng, Yuan, Chen, Guo, Su, Wang, Zhao, Zhang, Liu, Yan, Wu, Shi, Ye, Yu, Dong, Zhang, Ma, Pan, Gong, Liu, Ma, Wei, Cao, Huang, Jiang, Gao, Xiong, He, Huang, Xu, Wu, He, Wei, Jia, Wu, Xu, Zu, Zhou, Pan, Charles, Li, Hu, Liu, Chen, Wang, Liu, Qin, Liu, Yang, Bao, Du, Wu, Wang, Zhou, Wang, Li, Zhu, Zhang, Wang, Yang, Huang, Huang, Xu, Yang, and Lin]{kimiteam2025kimik15}
Kimi Team, Angang Du, Bofei Gao, Bowei Xing, Changjiu Jiang, Cheng Chen, Cheng Li, Chenjun Xiao, Chenzhuang Du, Chonghua Liao, Chuning Tang, Congcong Wang, Dehao Zhang, Enming Yuan, Enzhe Lu, Fengxiang Tang, Flood Sung, Guangda Wei, Guokun Lai, Haiqing Guo, Han Zhu, Hao Ding, Hao Hu, Hao Yang, Hao Zhang, Haotian Yao, Haotian Zhao, Haoyu Lu, Haoze Li, Haozhen Yu, Hongcheng Gao, Huabin Zheng, Huan Yuan, Jia Chen, Jianhang Guo, Jianlin Su, Jianzhou Wang, Jie Zhao, Jin Zhang, Jingyuan Liu, Junjie Yan, Junyan Wu, Lidong Shi, Ling Ye, Longhui Yu, Mengnan Dong, Neo Zhang, Ningchen Ma, Qiwei Pan, Qucheng Gong, Shaowei Liu, Shengling Ma, Shupeng Wei, Sihan Cao, Siying Huang, Tao Jiang, Weihao Gao, Weimin Xiong, Weiran He, Weixiao Huang, Weixin Xu, Wenhao Wu, Wenyang He, Xianghui Wei, Xianqing Jia, Xingzhe Wu, Xinran Xu, Xinxing Zu, Xinyu Zhou, Xuehai Pan, Y.~Charles, Yang Li, Yangyang Hu, Yangyang Liu, Yanru Chen, Yejie Wang, Yibo Liu, Yidao Qin, Yifeng Liu, Ying Yang, Yiping Bao, Yulun Du, Yuxin Wu, Yuzhi Wang, Zaida Zhou, Zhaoji Wang, Zhaowei Li, Zhen Zhu, Zheng Zhang, Zhexu Wang, Zhilin Yang, Zhiqi Huang, Zihao Huang, Ziyao Xu, Zonghan Yang, and Zongyu Lin.
\newblock Kimi k1.5: Scaling reinforcement learning with llms, 2025.
\newblock URL \url{https://arxiv.org/abs/2501.12599}.

\bibitem[Wei et~al.(2023)Wei, Zhao, Zhang, Zhu, Wang, Yang, Li, Cheng, Lü, Hu, Li, Yang, Luo, Wu, Liu, Cheng, Cheng, Zhang, Zhang, Lin, Wang, Ma, Dong, Sun, Chen, Peng, Liang, Yan, Fang, and Zhou]{skypile-150B}
Tianwen Wei, Liang Zhao, Lichang Zhang, Bo~Zhu, Lijie Wang, Haihua Yang, Biye Li, Cheng Cheng, Weiwei Lü, Rui Hu, Chenxia Li, Liu Yang, Xilin Luo, Xuejie Wu, Lunan Liu, Wenjun Cheng, Peng Cheng, Jianhao Zhang, Xiaoyu Zhang, Lei Lin, Xiaokun Wang, Yutuan Ma, Chuanhai Dong, Yanqi Sun, Yifu Chen, Yongyi Peng, Xiaojuan Liang, Shuicheng Yan, Han Fang, and Yahui Zhou.
\newblock Skywork: A more open bilingual foundation model, 2023.

\bibitem[Wen et~al.(2025)Wen, Cai, Xiao, He, An, Duan, Du, Liu, Tang, Lv, Zou, Deng, Jia, and Zhang]{wen2025light}
Liang Wen, Yunke Cai, Fenrui Xiao, Xin He, Qi~An, Zhenyu Duan, Yimin Du, Junchen Liu, Lifu Tang, Xiaowei Lv, Haosheng Zou, Yongchao Deng, Shousheng Jia, and Xiangzheng Zhang.
\newblock Light-r1: Curriculum sft, dpo and rl for long cot from scratch and beyond.
\newblock \emph{arXiv preprint arXiv:2503.10460}, 2025.

\bibitem[Yang et~al.(2025)Yang, Li, Yang, Zhang, Hui, Zheng, Yu, Gao, Huang, Lv, Zheng, Liu, Zhou, Huang, Hu, Ge, Wei, Lin, Tang, Yang, Tu, Zhang, Yang, Yang, Zhou, Zhou, Lin, Dang, Bao, Yang, Yu, Deng, Li, Xue, Li, Zhang, Wang, Zhu, Men, Gao, Liu, Luo, Li, Tang, Yin, Ren, Wang, Zhang, Ren, Fan, Su, Zhang, Zhang, Wan, Liu, Wang, Cui, Zhang, Zhou, and Qiu]{qwen3}
An~Yang, Anfeng Li, Baosong Yang, Beichen Zhang, Binyuan Hui, Bo~Zheng, Bowen Yu, Chang Gao, Chengen Huang, Chenxu Lv, Chujie Zheng, Dayiheng Liu, Fan Zhou, Fei Huang, Feng Hu, Hao Ge, Haoran Wei, Huan Lin, Jialong Tang, Jian Yang, Jianhong Tu, Jianwei Zhang, Jianxin Yang, Jiaxi Yang, Jing Zhou, Jingren Zhou, Junyang Lin, Kai Dang, Keqin Bao, Kexin Yang, Le~Yu, Lianghao Deng, Mei Li, Mingfeng Xue, Mingze Li, Pei Zhang, Peng Wang, Qin Zhu, Rui Men, Ruize Gao, Shixuan Liu, Shuang Luo, Tianhao Li, Tianyi Tang, Wenbiao Yin, Xingzhang Ren, Xinyu Wang, Xinyu Zhang, Xuancheng Ren, Yang Fan, Yang Su, Yichang Zhang, Yinger Zhang, Yu~Wan, Yuqiong Liu, Zekun Wang, Zeyu Cui, Zhenru Zhang, Zhipeng Zhou, and Zihan Qiu.
\newblock Qwen3 technical report.
\newblock \emph{arXiv preprint arXiv:2505.09388}, 2025.

\bibitem[Ye et~al.(2025)Ye, Huang, Xiao, Chern, Xia, and Liu]{ye2025limoreasoning}
Yixin Ye, Zhen Huang, Yang Xiao, Ethan Chern, Shijie Xia, and Pengfei Liu.
\newblock Limo: Less is more for reasoning, 2025.
\newblock URL \url{https://arxiv.org/abs/2502.03387}.

\bibitem[Yu et~al.(2025)Yu, Zhang, Zhu, Yuan, Zuo, Yue, Dai, Fan, Liu, Liu, Liu, Lin, Lin, Ma, Sheng, Tong, Zhang, Zhang, Zhang, Zhu, Zhu, Chen, Chen, Wang, Yu, Song, Wei, Zhou, Liu, Ma, Zhang, Yan, Qiao, Wu, and Wang]{yu2025dapoopensourcellmreinforcement}
Qiying Yu, Zheng Zhang, Ruofei Zhu, Yufeng Yuan, Xiaochen Zuo, Yu~Yue, Weinan Dai, Tiantian Fan, Gaohong Liu, Lingjun Liu, Xin Liu, Haibin Lin, Zhiqi Lin, Bole Ma, Guangming Sheng, Yuxuan Tong, Chi Zhang, Mofan Zhang, Wang Zhang, Hang Zhu, Jinhua Zhu, Jiaze Chen, Jiangjie Chen, Chengyi Wang, Hongli Yu, Yuxuan Song, Xiangpeng Wei, Hao Zhou, Jingjing Liu, Wei-Ying Ma, Ya-Qin Zhang, Lin Yan, Mu~Qiao, Yonghui Wu, and Mingxuan Wang.
\newblock Dapo: An open-source llm reinforcement learning system at scale, 2025.
\newblock URL \url{https://arxiv.org/abs/2503.14476}.

\bibitem[Yue et~al.(2025)Yue, Yuan, Yu, Zuo, Zhu, Xu, Chen, Wang, Fan, Du, Wei, Yu, Liu, Liu, Liu, Lin, Lin, Ma, Zhang, Zhang, Zhang, Zhu, Zhang, Liu, Wang, Wu, and Yan]{yue2025vapoefficientreliablereinforcement}
Yu~Yue, Yufeng Yuan, Qiying Yu, Xiaochen Zuo, Ruofei Zhu, Wenyuan Xu, Jiaze Chen, Chengyi Wang, TianTian Fan, Zhengyin Du, Xiangpeng Wei, Xiangyu Yu, Gaohong Liu, Juncai Liu, Lingjun Liu, Haibin Lin, Zhiqi Lin, Bole Ma, Chi Zhang, Mofan Zhang, Wang Zhang, Hang Zhu, Ru~Zhang, Xin Liu, Mingxuan Wang, Yonghui Wu, and Lin Yan.
\newblock Vapo: Efficient and reliable reinforcement learning for advanced reasoning tasks, 2025.
\newblock URL \url{https://arxiv.org/abs/2504.05118}.

\bibitem[Zhu et~al.(2024)Zhu, Li, Wen, and Guo]{zhu-CFLUE-2024-benchmarking}
Jie Zhu, Junhui Li, Yalong Wen, and Lifan Guo.
\newblock Benchmarking large language models on cflue - a chinese financial language understanding evaluation dataset.
\newblock In \emph{Findings of ACL}, pp.\  5673--5693, 2024.

\bibitem[Zhu et~al.(2025)Zhu, Chen, Dou, Li, Guo, Chen, and Zhang]{dianjin-r1}
Jie Zhu, Qian Chen, Huaixia Dou, Junhui Li, Lifan Guo, Feng Chen, and Chi Zhang.
\newblock Dianjin-r1: Evaluating and enhancing financial reasoning in large language models.
\newblock \emph{arxiv.org/abs/2504.15716}, 2025.

\end{thebibliography}
\bibliographystyle{iclr2025_conference}

\clearpage
\appendix
\section{Appendix}

\subsection{Training Datasets}

\begin{table}[htb]
    \centering
    \small
    \begin{tabular}{
    >{\Centering\arraybackslash}p{22ex}
    >{\Centering\arraybackslash}p{10ex}
    >{\Centering\arraybackslash}p{15ex}
    >{\Centering\arraybackslash}p{10ex} 
    >{\RaggedRight\arraybackslash}p{25ex}}
        \toprule
        {Name} & {Domain} & {Source} & {Size} & {Notes} \\ 
        \midrule
        \rowcolor{mygray} \multicolumn{5}{c}{CPT Data} \\
        \midrule
        Acc Learn & Finance & In-house & 329,419 (tokens) & Content from registered accountant textbooks \\
        FinCorpus QA & Finance & Open-source & 23,283,970 (tokens) & Duxiaoman open-source financial dataset \\
        Industry Corpus Finance & Finance & Open-source & 24M (tokens) & BAAI open-source financial dataset \\
        InfiMM-WebMath-Edu-zh & General & Open-source & 47M (tokens) & Open-source Chinese mathematics pretrain dataset \\
        SkyPile-150B & General & Open-source & 47M (tokens) & Open-source Chinese general pretrain dataset \\
        Dolma & General & Open-source & 47M (tokens) & Open-source English general pretrain dataset, including maths and code \\
        \midrule
        \rowcolor{mygray} \multicolumn{5}{c}{SFT Data} \\
        \midrule
        Duxiaoman FinQA & Finance & Open-source & 10K (items) & Duxiaoman open-source financial dataset \\
        Fin CPA Agent-instruct & Finance & Synthetic & 1121 (items) & Synthesized using agent-instruct, based on in-house Fin CPA data\\
        CFLUE & Finance & Open-source & 20K (items) & Chinese Financial Language Understanding Evaluation \\
        Chinese-DeepSeek-R1-Distill-data-110K & Finance & Open-source & 30K (items) & Chinese-Data-Distill-From-R1, only using mathematics portion with scores of 10 \\
        \midrule
        \rowcolor{mygray} \multicolumn{5}{c}{RL Data} \\
        \midrule
        Fin CPA & Finance & In-house & 215 (items) & Real-world data from accountant exams \\
        Duxiaoman Fill-in-blanks & Finance & Open-source & 23K (items) & Extracted from Duxiaoman FinQA (no overlap with SFT data) then converted to open-ended questions \\
        Dianjin Fill-in-blanks & Finance & Open-source & 2K (items) & Extracted from CFLUE (no overlap with SFT data) then converted to open-ended questions and only keeping entries with answer length $< 15$ \\
        AM-Math-Difficulty-RL \citep{2025diff-aware-rl} & General & Open-source & 4K (items) & Math dataset with three difficulty levels. We only extract a portion of the overall dataset: 400 easy, 600 intermediate, and 3000 hard. \\
        \bottomrule
    \end{tabular}
    \caption{Detailed info for training dataset}
    \label{tab:train-dataset}
\end{table}

\clearpage
\subsection{Evaluation Datasets}

\begin{table}[htbp]
    \centering
    \begin{tabular}{
    >{\Centering\arraybackslash}p{15ex} 
    >{\Centering\arraybackslash}p{15ex} 
    >{\Centering\arraybackslash}p{6ex} 
    >{\RaggedRight\arraybackslash}p{40ex}}
        \toprule
        {Name} & {Source} & {Size} & {Notes} \\
        \midrule
        Fin CPA & In-house & 215 & Gathered from in-house financial tasks \\
        Fin CCR & Open-source & 602 & Reasoning \& calculation questions selected from Fin-Eva and OpenFinData \\
        Finance IQ & Open-source & 7173 & 
        Duxiaoman open-source finance evaluation dataset
        \\
        Fine-Eva & Open-source & 3584 & 
        SUFE Open source evaluation dataset, contains knowledge and reasoning questions
        \\
        CFLUE & Open-source & 3864 & 
        Chinese Financial Language Understanding Evaluation (CFLUE), a comprehensive evaluation for LLM's ability in the Chinese financial domain
        \\
        OpenFinData & Open-source & 650 & An open-source financial evaluation dataset curated by East Money and Shanghai AI Lab \\
        Math 500 & Open-source & 500 & General evaluation benchmark for mathematical performance  \\
        \bottomrule
    \end{tabular}

    \caption{The benchmarks used for evaluation}
    \label{tab:benchmark_info}
\end{table}

\subsection{RL optimization algorithms} \label{vapo-explain}

\begin{figure}[htbp]
\begin{center}
\includegraphics[width=0.8\textwidth]{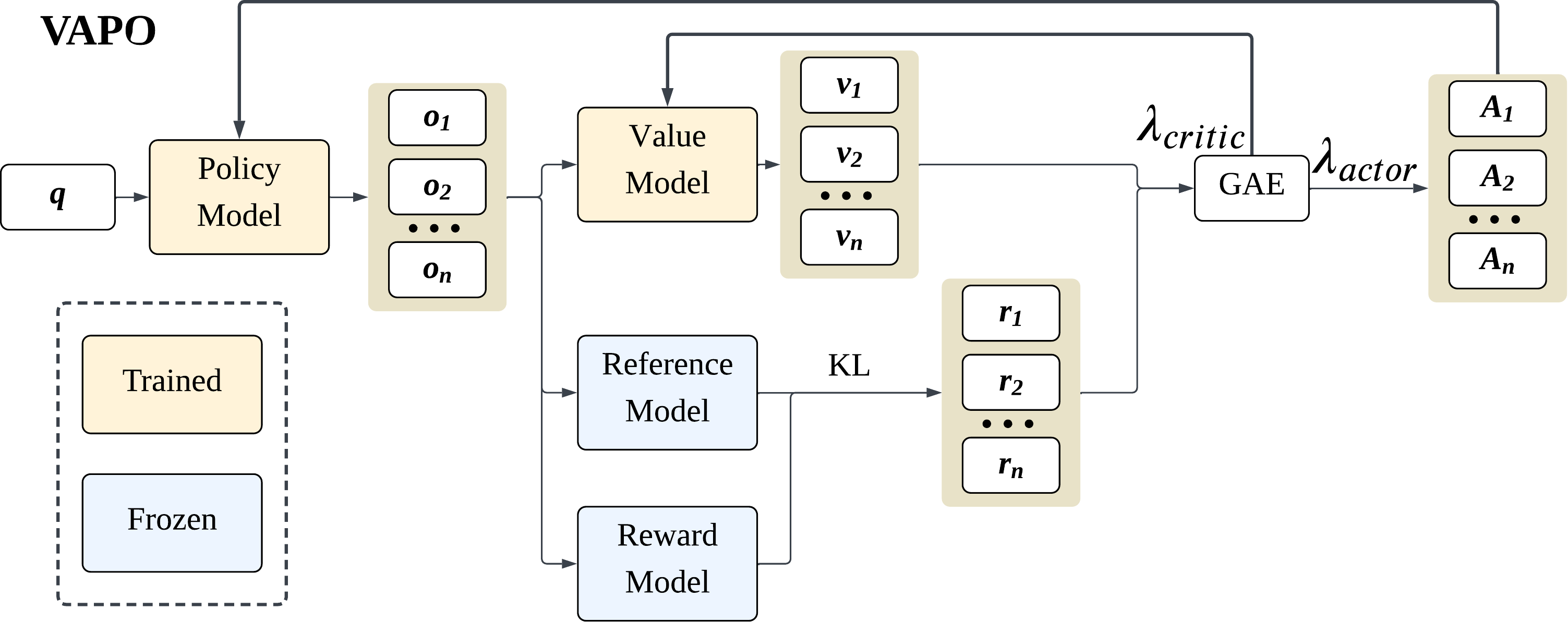}
\end{center}
\caption{RL training process using VAPO algorithm}
\label{fig:vapo-diagram}
\end{figure}

In our experiments, we predominantly use the VAPO algorithm for RL. VAPO utilizes two losses: PPO \citep{schulman2017ppo} loss and a positive LM loss. Formally, it computes the policy gradient loss as follows:

\begin{align}    
    \mathcal{L}_{\text{PPO}} (\theta) & = - \frac{1}{\sum_{i=1}^{G}|o_i|}\sum_{i=1}^G \sum_{t}^{{o_i}} \min(r_{i, t}(\theta) \hat{A}_{i, t}, \text{clip}\left(r_{i, t}(\theta), 1 - \epsilon_{low}, 1 + \epsilon_{high})\hat{A}_{i, t} \right) \\
    \mathcal{L}_{\text{NLL}} (\theta) & = - \frac{1}{\sum_{o_i \in \mathcal{T} } |o_i|} \sum_{o_i \in \mathcal{T}} \sum_{t=1}^{|o_i|}\log \pi_\theta (a_t | s_t) \\
    \mathcal{L}_{\text{VAPO}} (\theta) & = \mathcal{L}_{\text{PPO}} (\theta) + \mu \cdot \mathcal{L}_{\text{NLL}} (\theta)
\end{align}

where $\pi_\theta(a|s)$ is the policy parameterized by $\theta$, $\pi_{\theta_{old}}(a|s)$ is the old policy from the previous iteration, $r_\theta $ is the probability ratio $\frac{\pi_\theta (a_t | s_t)}{\pi_{\theta_{old}} (a_t | s_t)}$, \textit{G} is the size of training batch, $o_i$ is the trajectory/output of the \textit{i}th sample; $\hat{A}$ denotes the estimated advantage at time $t$, $\mathcal{T}$ denotes the set of responses that are correct, $\mu$ is a weighting coefficient for the negative log-likilihood (NLL) loss and $\epsilon_{low}, \epsilon_{high}$ denote hyperparameters that control the clipping range that limit the policy change during each update step.

For advantage calculations, VAPO uses a length-adaptive coefficient $\lambda$ in the GAE \citep{schulman2018gae} formula. For an output with length $l$, the corresponding $\lambda_{actor}$ as well as the advantage estimate $\hat{A}_t$ at each time step are calculated as follows:

\begin{align}
    \lambda_{actor} & = 1 - \frac{1}{\alpha l} \\
    \hat{A}_t & = \sum_{k=0}^{l-t-1} (\gamma \lambda_{actor})^{k} \left( R(s_t, a_t) + \gamma V(s_{t+1} - V(s_t) \right)
\end{align}

\subsection{Sample responses} \label{sample-comparison}

\begin{figure}[htbp]
\begin{center}
\includegraphics[width=\textwidth]{svg-inkscape/response-comparison_svg-raw.pdf}
\end{center}
\caption{Comparison of a sample response for R32B and R32B-0}
\label{fig:sample-comparison-fig}
\end{figure}

\end{document}